\pdfoutput=1

\documentclass[11pt]{article}

\usepackage{ACL2023}
\usepackage{times}
\usepackage{latexsym}
\usepackage{array}
\usepackage{booktabs}
\usepackage{multirow}
\usepackage{changepage}
\usepackage{multicol}
\usepackage{amsmath}
\usepackage{tabularx}
\usepackage{hyperref}
\usepackage{adjustbox}
\usepackage{tikz}
\usepackage{tcolorbox}

\usepackage[T1]{fontenc}

\usepackage[utf8]{inputenc}

\usepackage{microtype}
\usepackage{tcolorbox}

\usepackage{inconsolata}

\title{ 
Leveraging LLM For Synchronizing Information Across Multilingual Tables}

\author{
Siddharth Khincha\textsuperscript{\rm 1},
Tushar Kataria\textsuperscript{\rm 2}
Ankita Anand\textsuperscript{\rm 1},
\textbf{Dan Roth}\textsuperscript{\rm 3},
\textbf{Vivek Gupta}\textsuperscript{\rm 4}\thanks{~~Corresponding Author}~\\ 
\textsuperscript{\rm 1}IIT Guwahati,
\textsuperscript{\rm 2}University of Utah \\
\textsuperscript{\rm 3}University of Pennsylvania,
\textsuperscript{\rm 4}Arizona State University\\
\small \{s.khincha,ankita.anand\}@iitg.ac.in, 
 tkataria@cs.utah.edu, danroth@seas.upenn.edu, vgupt140@asu.edu \\
}

\begin{document}
\maketitle
\begin{abstract}

The vast amount of online information today poses challenges for non-English speakers, as much of it is concentrated in high-resource languages such as English and French. Wikipedia reflects this imbalance, with content in low-resource languages frequently outdated or incomplete. Recent research has sought to improve cross-language synchronization of Wikipedia tables using rule-based methods. These approaches can be effective, but they struggle with complexity and generalization. This paper explores large language models (LLMs) for multilingual information synchronization, using zero-shot prompting as a scalable solution. We introduce the \textit{Information Updation} dataset, simulating the real-world process of updating outdated Wikipedia tables, and evaluate LLM performance. Our findings reveal that single-prompt approaches often produce suboptimal results, prompting us to introduce a task decomposition strategy that enhances coherence and accuracy. Our proposed method outperforms existing baselines, particularly in Information Updation (1.79\%) and Information Addition 
, highlighting the model’s strength in dynamically updating and enriching data across architectures. 
\end{abstract}

\section{Introduction}

In today's digital era, nearly every subject/domain is discoverable online. With global access to high-speed internet expanding, the volume of information grows exponentially\footnote{\href{https://www.statista.com/topics/1145/internet-usage-worldwide/}{Statista Worldwide Internet Usage}}\footnote{\href{https://www.cisco.com/c/en/us/solutions/collateral/executive-perspectives/annual-internet-report/white-paper-c11-741490.html}{Cisco Annual Internet Report}}. From movies and celebrities to elections and corporate news, a vast array of topics is just a click away for those with access. However, since developed countries—particularly English-speaking ones—were early adopters of the internet, much online content is tailored to English-speaking audiences\footnote{\href{https://en.wikipedia.org/wiki/Wikipedia:List_of_Wikipedians_by_number_of_edits}{Wikipedia: List of Wikipedians by Number of Edits}}. This is evident on platforms such as Wikipedia and YouTube, where English dominates\footnote{\href{https://en.wikipedia.org/wiki/List_of_Wikipedias}{Wikipedia: Active Users by Language}}. Although the number of non-English users is growing, underrepresented languages such as Afrikaans, Cebuano, and Hindi still face a significant information gap \cite{10.1145/2207676.2208553}.

\begin{figure*}
    \centering
    \includegraphics[width=0.8\linewidth]{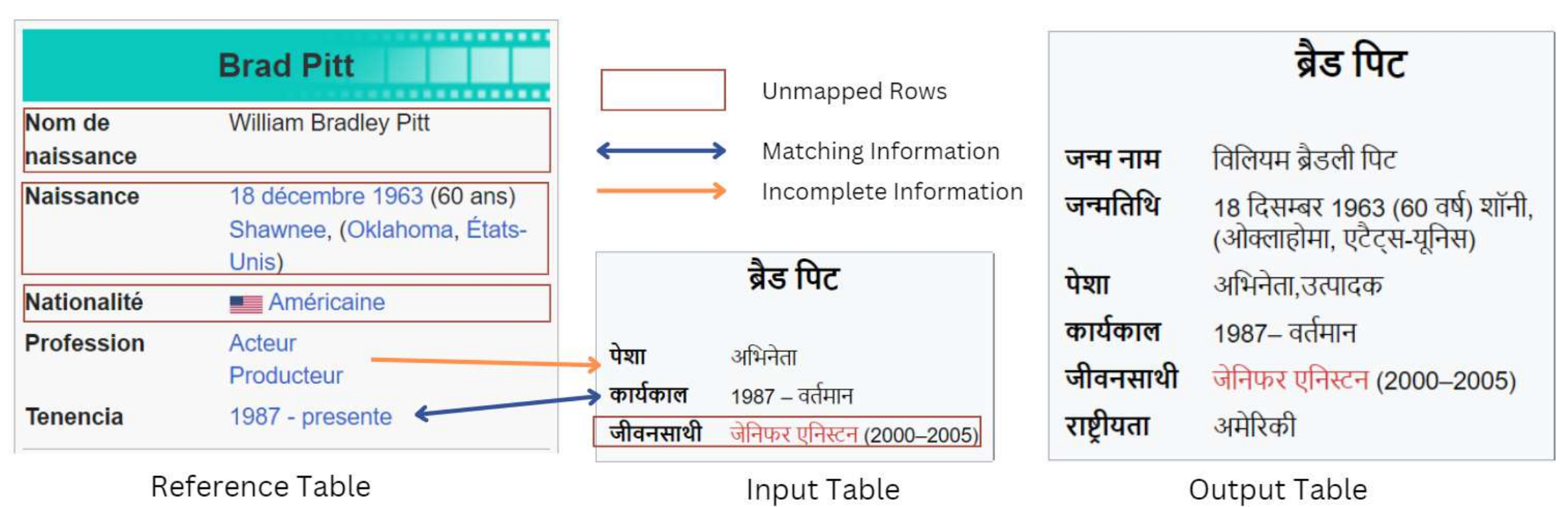}
    \vspace{-0.5em}
    \caption{Example of information synchronization across multilingual tables. A reference table in a high-resource language is used to update outdated input tables in a low-resource language, resulting in an updated output table in the low-resource language.}
    \label{fig:examples}
    \vspace{-1.5em}
\end{figure*}

As shown by \citet{khincha-etal-2023-infosync} in their case study on Wikipedia's entity-centric tables, information in Wikipedia infoboxes \cite{zhang2020web} is heavily skewed toward high-resource languages such as English, Spanish, and French. They found that tables in low-resource languages often lack key information and are frequently outdated or inaccurate \cite{jang2016utilization,nguyen2018automatically}. This disparity is especially concerning in the digital age, where misinformation on widely accessible platforms can have far-reaching consequences. To address this, \citet{khincha-etal-2023-infosync} developed the {\sc InfoSync} dataset to analyze information synchronization issues across 14 languages. They proposed aligning tables by matching similar keys and using rule-based methods to transfer and update information in tables across languages. Figure \ref{fig:examples} illustrates an example of information synchronization across multilingual tables (Spanish and Hindi). However, {\sc InfoSync}'s approach has a key limitation: rule-based methods become increasingly complex as new corner cases emerge, making generalization challenging. A more effective alternative is leveraging current large language models (LLMs) for zero-shot prompting, providing an easily scalable solution for these tasks.

With new and advanced LLMs (such as GPT, Mitral, LLAMA \cite{brown2020language,touvron2023llama,achiam2023gpt,jiang2023mistral}) being released every year, the zero-shot prompting capabilities of these models are improving with each new training iteration. These LLMs are consistently approaching, and in some cases surpassing, human performance across various NLP applications \cite{achiam2023gpt}. LLMs excel at text-generation \cite{achiam2023gpt}, text modification based on provided prompts \cite{raffel2020exploring}, and text refinement by correcting errors \cite{davis2024prompting,liu2024proofread,li2024rethinking}. Given the substantial advancements in LLM capabilities, this paper poses the following questions:  \textit{Can LLMs be leveraged for information synchronization in multilingual Wikipedia entity-centric tables? If so, how can LLMs be utilized, and how effective are they compared to the rule-based methods proposed by \citet{khincha-etal-2023-infosync}?}

To investigate these questions, we first construct an \textit{Information Updation} dataset. This dataset simulates the information updating task by using outdated versions of tables to represent old information, and comparing them with the latest versions that have been manually corrected to address any missing or outdated content. Additionally, we assess the performance of state-of-the-art large language models (LLMs), specifically GPT-4 \cite{achiam2023gpt}, for the task of information synchronization in entity-centric tables across multiple languages. To evaluate performance on the \textit{Information Updation} task, we propose novel automated metrics that offer valuable insights into model performance and identify potential areas for improvement in future iterations. 

Our initial experiments reveal that relying on a single prompt for multilingual information synchronization with LLMs yields suboptimal performance, frequently producing incoherent edits. To enhance these results, we propose a task decomposition approach. Our proposed method uses multiple prompts to address smaller, more manageable subtasks, which are then connected in a sequential pipeline to generate the final output. Task decomposition has shown promise in improving performance across a range of complex NLP applications \cite{khot2022decomposed,ma2024task,wang2024hierarchical}, and we find it similarly beneficial in our work. 
Our work makes the following contributions:
\begin{itemize}
\setlength\itemsep{-0.5em} 
    \item We create an \textit{Information Updation} dataset by sampling older versions of the same Wikipedia pages. This dataset simulates the real world process of updating Wikipedia infoboxes with human input, reflecting the task of correcting and adding new information over time.
    \item We employ large language models (LLMs) for zero-shot automated multilingual information synchronization in entity-centric Wikipedia tables. By utilizing prompt-based task decomposition, we significantly enhance the accuracy and coherence of the results.
    \item Develop novel evaluation metrics for the Information Updation task, alongside conducting a thorough error analysis to identify the limitations of current state-of-the-art LLM models.
\end{itemize}

Code and dataset are available at \href{https://zero-shot-llm-infosync.github.io/zero-shot-llm-infosync/}{https://zero-shot-llm-infosync.github.io/zero-shot-llm-infosync/}.


\section{Proposed Methodology}
Information synchronization for Wikipedia infoboxes involves updating outdated rows in the table by conditionally modifying attributes, values, or both, using data from reference infoboxes(which have updated information).  Consider a source table ($T_S$) in language ($L_i$) that contains missing or outdated information. We also have a reference table ($T_R$) in language ($L_j$), which provides the missing and updated information not found in $T_S$. Additionally, we assume access to gold-standard updated table ($T_G$) in language ($L_i$), which can be regarded as having been manually curated.

The \textit{Information Synchronization} task is to update the table $T_S$ using only the information available in $T_S$ and $T_R$, with the goal of matching the updated table to $T_G$. Previous work by \citet{khincha-etal-2023-infosync} approaches this problem with a two-step methodology: (i) \textit{Information Alignment}, which involves identifying similar rows across different tables using cosine similarity, and (ii) \textit{Information Update}, which utilizes rule-based methods to update $T_S$. In contrast, we propose to solve this task using large language models (LLMs) to provide a more automated and sophisticated solution, bypassing the need for elementary similarity measures and rule-based methods. We propose solving the information synchronization problem using various prompts, as outlined below: 

\textbf{Simple Prompt.} With large-scale pre-training of language modeling, new language models such as GPT4, LLaMA, and Gemini Pro support prompt-based (instruction set) evaluations, which do not require any finetuning. As a baseline prompt, we explain the task of information synchronization in the prompt giving details of types of missing information that might be presented between the two tables such as outdated information, missing information, or inconsistent information. The model is tasked to create an output table that has updated information from both source and reference tables. Here, we ask the model to give more importance to reference table information while creating an updated output table. 

\textbf{Elementary Task decomposition within a Single Prompt}: To test whether a single prompt can give reasonable outputs even when directed to do task decomposition as an intermediate step, we propose 
\textit{Align-Update Decomposition} prompt. In this prompt, the model is instructed to first implicitly align all corresponding information between the two tables. Once these alignments are automatically generated, the model should carefully review each alignment to identify and remove any outdated information wherever necessary. Additionally, the model is explicitly instructed to add missing rows that could not be mapped during the alignment process. This prompt is inspired by the task decomposition of \citet{khincha-etal-2023-infosync}, which does the same with rule-based approach.

\textbf{Hierarchical Task Decomposition Prompt}. Instead of creating a single instruction set for the task of Information synchronization. We do a hierarchical decomposition of the task and create prompts for each step. These prompts are applied sequentially, with the output of the last prompt as input to the next prompt in the hierarchy. Different hierarchical steps for this prompt are:
\setlength\itemsep{-0.25em} 
\begin{itemize}
\setlength\itemsep{-0.25em} 
    \item \textbf{Translation}: All tables ($T_S$, $T_R$, and $T_G$) from different languages are converted to English. English is selected as the base language because most state-of-the-art LLMs are largely trained on curated English data, resulting in higher accuracy for complex reasoning and analysis tasks performed in English compared to other languages.
    \item \textbf{Knowledge graphs conversion}: The translated source and reference tables are then converted into knowledge graphs. Our experiments indicate that the subsequent hierarchical steps are more effective when using knowledge graphs rather than infoboxes/tables. LLMs perform better reasoning over knowledge graphs. 
    \item \textbf{Merging or alignment}: The source($KG_S$) and reference($KG_R$) knowledge graphs are merged to create a unified knowledge graph that consolidates all the information from both sources. Merging of knowledge graph is equal to alignment step described in the section above. This merging process helps eliminate redundant information from unresolved conflicts and enhances the inclusion of missing details. During the merge step, the model first gathers all necessary information, and in the subsequent update step, it makes the relevant adjustments.
    \item \textbf{Update:} The merged knowledge graphs is used to update information in the source knowledge graphs. Due to these being node operations, these are fast and have better interpretablity.
\end{itemize}

After the update step, the revised knowledge graphs are converted back into tables. These tables are then translated back into the original languages of the source tables. We compare these three prompt designs for the task of information synchronization with relevant ablations for heirarchical task decomposition. Prompt examples are shown in Appendix \ref{sec:Prompts}.

\section{{\sc InfoUpdate} Benchmark}
\citet{khincha-etal-2023-infosync} concentrated on developing a dataset for information alignment tasks, i.e., aligning similar keys across tables coming from different languages. They employed a rule-based method for updates and conducted human evaluations based on these updates to recommend edits on Wikipedia pages. However, they did not create a dataset or propose an automated method for evaluating approaches to information updation.  To address this, we introduce a new human-annotated dataset ({\sc InfoUpdate}) for the \textit{Information Updation} task focused on Wikipedia infoboxes. This dataset comprises approximately 950 annotated instances across 9 categories: Album, Athlete, City, College, Company, Country, Musician, Person, and Stadium, spanning 14 languages including Spanish, French, German, Arabic, Hindi, Korean, Russian, Afrikaans, Cebuano, Swedish, Dutch, Turkish, and Chinese. Additional dataset statistics are shown in Appendix Table \ref{tab:combined_tables}.

\textbf{{\sc InfoUpdate} Construction.} 
We construct the dataset by extracting two versions of the same Wikipedia table entity from different time periods. For a table in Category \textit{C} titled \textit{T} and language($L_i$), we extract the \textit{Old} version from 2018 (source table) and the \textit{New} version from 2023 (current at the time of extraction). The new version of the table is extracted from two or more different languages, where the table in the same language serves as the target, and the table in a different language acts as a reference (or additional information), i.e., updated table. This setup is designed to simulate a real-world implementation of the Information Synchronization task for entity centric semistructured tables. For every entity in a single instance of the task, the dataset contains 3 tables source table $T_S$, reference table $T_R$ and a gold table $T_G$. 
\begin{itemize}
\setlength\itemsep{-0.25em}
    \item The \textbf{Source Table} $T_S$ is the \textit(Old) outdated version of the entity in language $L_i$.
    \item The \textbf{Reference Table} $T_R$ is the \textit{New} updated version  of the entity in language $L_j$($i$ $\neq$ $j$)
    \item The \textbf{Gold Table} $T_G$ is the human annotated version, which is manually created by synchronising the \textit{New} updated versions of the entity in the languages $L_i$ and $L_j$.
\end{itemize}
The information updation task is to update rows in source table ($T_S$) using reference table ($T_R$) as context information, so that the resulting generated table, referred to as the output table ($T_O$), has the same information as Gold Table ($T_G$).

\textbf{{\sc InfoUpdate} Verification.} Human annotators are tasked to ensure two aspects:(a) The gold table contains the complete information present in both the input and reference table combined without any redundancy. (b) The gold table is consistent, resolving all conflicts or missing data without adding new information not found in the Source or Reference tables.

\section{{\sc InfoUpdate} Evaluation Metric} 
\label{sec:eval_metric}

For evaluation, we first experimented with an approach similar to \citet{chiang-lee-2023-large}, where we tasked the model with comparing the output table against a reference table and providing a score. However, this approach proved to be highly unstable, resulting in vastly different responses even in low-temperature settings. Moreover, the model often overlooked multiple rows during evaluation, making it difficult to establish a consistent scoring system due to the subjective nature of the process.
Therefore, we propose our novel evaluation metric consisting of two main steps: (a) \textit{Information Alignment Evaluation}  and (b) \textit{Information Update Evaluation}, aimed at addressing the aforementioned issues.

\begin{figure}[!htb]
    \centering
    \includegraphics[width=0.70\linewidth]{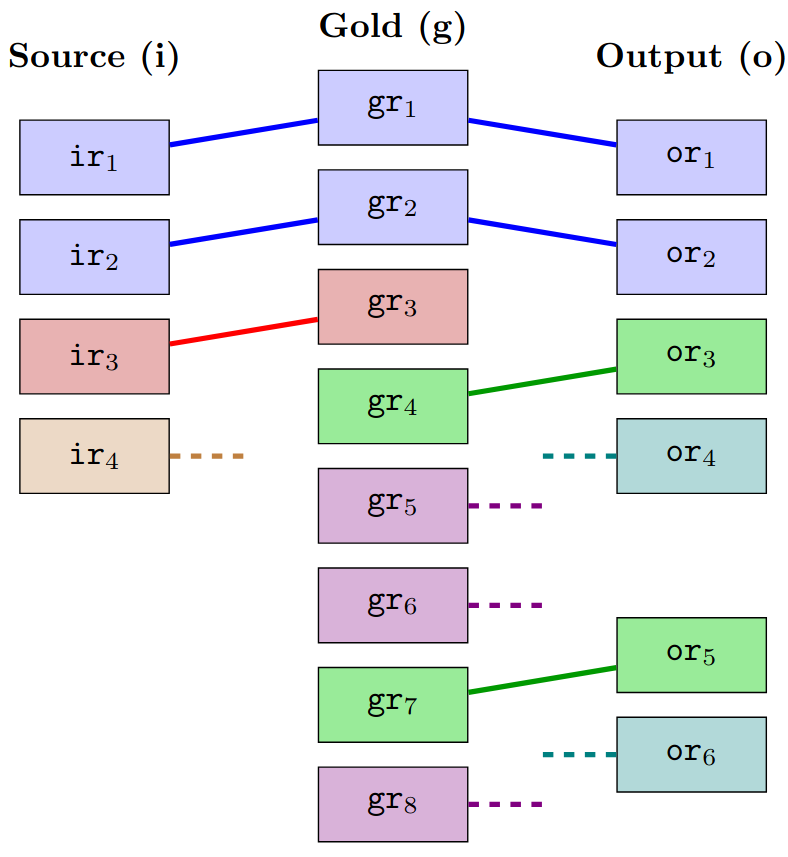}
     \vspace{-0.5em}
    \caption{\textbf{Alignment Groups For Information Alignment}. All rows highlighted in blue and connected by blue lines in the Source, Gold, and Output tables are tri-aligned, meaning they contain the same information across all three tables. Rows highlighted in red or green are bi-aligned, indicating that the information is consistent either between the Input and Gold tables or the Gold and Output tables. The remaining rows are unaligned, containing differing information. }
    \label{fig:alignment_diagram}
    \vspace{-1.em}
\end{figure}

\paragraph{Information Alignment Evaluation.} In this step, we first create a mapping of similar information or alignments between \{{\sc Source Table}($T_S$), {\sc Gold Table}($T_G$)\} and \{{\sc Output Table (Generated Table)}($T_O$), {\sc Gold Table}($T_G$)\} tables. {\sc Reference table} ($T_R$) is not used during the evaluation process, as it in different language and only used as referenced for updating source ($T_S$) tables. 

Alignment between ($T_S$) and ($T_G$) gives us a metric of \emph{information already present in the source when compared to gold}, whereas alignment between ($T_O$), and ($T_G$) informs us of \emph{extra alignments added due to the information generated by the synchronization model, i.e., table updation}. The alignments are compared and separated into three groups: Tri-Aligned, Bi-Aligned, and Un-Aligned. Figure ~\ref{fig:alignment_diagram} shows pictorial the meaning and different between the three alignment group we defined and Table \ref{tab:components-associations} shows the results for each types of alignments. Formal definitions in logic statements can be found in Appendix Table \ref{tab:set-theoretic-definitions}.
    
\noindent \textbf{- Tri-Aligned}: These refer to the table keys that are common across all three tables: the {\sc Source, Output}, and {\sc Gold}. $(ir_1,gr_1,or_1)$ and $(ir_2,gr_2,or_2)$ are trialigned rows in example Figure \ref{fig:alignment_diagram}. These represent information that is either kept intact by the model from source to output, or can also have cases where the information was incomplete in source table but was completed by model operations. 

\noindent \textbf{- Bi-aligned}: When the table keys are common across pairs of tables {\sc Gold-Output} $(gr_4,or_3),(gr_7,or_5)$ or {\sc Gold-Source} $(ir_3,gr_3)$, but not across all three tables, these define Bi-aligned rows. The number of Gold elements aligned with the Output but not the Input indicates the amount of information added, which was not present before $(gr_4,or_3),(gr_7,or_5)$, while the number of Gold elements aligned with the Input but not the Output represents the amount of relevant information deleted $(ir_3,gr_3)$.

\noindent \textbf{- Un-aligned}: These are keys remaining in {\sc Source, Output} and {\sc Gold} tables after tri- and bi-aligned keys are removed from tables. We have three types of unaligned rows as follows: 

(a.) \emph{Unaligned {\sc Source Table} keys} ($ir_4$), refers to redundant input information deleted in the output table. 

(b.) \emph{Unaligned {\sc Output table} keys} $(or_4,or_6)$,  refers to hallucinated/noisy/irrelevant information added to output table not present in either (Source or Gold).

(c.) \emph{Unaligned {\sc Gold table} keys} $(gr_5,gr_6,gr_8)$,  refer to information gaps that the model could not add to the Output table, either due to model inaccuracies or because the information is missing in the Source and Reference tables. 

\paragraph{Information Updation Evaluation.} We evaluate each alignment pair from both {\sc Source-Gold} alignments and {\sc Output-Gold} alignments for semantic equivalence using Large Language Models (Eval$_{\text{LLM}}$). Here, we check if the align information is fully-matching, partially matching or contradictory to each other. 
For each aligned key-value pair, the LLM is instructed to examine the information, translate it into English, and decompose it into fine-grained atomic details. These atomic details are then categorized into four distinct groups: (a) \textbf{Similar and Consistent (SCT)}—information appearing in both tables with consistent values; (b) \textbf{Similar and Contradictory (SCD)}—information present in both tables but exhibiting contradictory or conflicting values; (c) \textbf{Table 1 Unique (T1U)}—information unique to Table 1, not found in Table 2; and (d) \textbf{Table 2 Unique (T2U)}—information unique to Table 2, not present in Table 1. These four categories are used to calculate the precision and recall for evaluation as follows:

\vspace{-1.35em}
{\small {
\[
\text{Precision} = \frac{| \text{SCT} |}{| \text{SCT} | + | \text{SCD} | + | \text{T1U} |}
\]

\[
\text{Recall} = \frac{| \text{SCT} |}{| \text{SCT} | + | \text{SCD} | + | \text{T2U} |}
\]

\[
\text{F1 Score} = \frac{2 \times \text{Precision} \times \text{Recall}}{\text{Precision} + \text{Recall}}
\]
}}

Here, | X | denotes the cardinality of the set X, indicating the number of elements within set X. The precision and recall scores are normalized by dividing them by the length of the gold table ($T_G$), ensuring a fair comparison. We use these measures to evaluate \textbf{SCT}, \textbf{SCD}, \textbf{T1U} and \textbf{T2U} for different models.
We used this metric for all information alignment. The detail metric for each information alignment types (tri, bi, and un) is shown in Appendix Table \ref{tab:notation-metrics-details}, grounded with the running examples shown in the Figure \ref{fig:alignment_diagram}.


 \section{Experiments and Results}

\paragraph{Alignment Models.} For \textit{information alignment}, we employ an ensemble of multiround voting methods that combine InfoSync—a deterministic alignment algorithm—with LLM-based alignment utilizing few-shot chain-of-thought prompting with detailed instructions. We conducted three runs each with GPT-3.5 and Gemini 1.5 Flash Pro, followed by majority voting to establish the LLM alignment. The majority-voted alignments from both GPT-3.5 and Gemini 1.5 Flash Pro are then integrated into InfoSync, using majority voting again to achieve the final alignment. We refer to this approach as the multi-voting scheme. 

\paragraph{Updation Models.} For \textit{information updation}, we utilize three large language models (LLMs) for our experiments: GPT-3.5, Gemini 1.5 Flash Pro  \cite{reid2024gemini}, and LLAMA 3.0 (70B)  \cite{llama3modelcard}. The first two are closed-source models accessible only via API, whereas the latter is an open-source model. These models were selected as they represent state-of-the-art performance or are at least close to it and fall within the computational budget of our project. We believe that our results will also be applicable to other closed and open-source LLMs.

\paragraph{Evaluation Strategy.} Evaluating the efficacy of large language models (LLMs) involves two key steps: (a) information alignment, which prepares the tables for comparison, and (b) information evaluation, which assesses the semantics. We evaluate both of these as following: 

\textit{1. Alignment Evaluation.} The alignments are compared with human alignments for both {\sc Input-Output} (200 pairs) and {\sc Output-Gold} (200 pairs), for GPT 3.5 generated updated tables with our decomposition approach.

\textit{2. Updation Evaluation.}  Human evaluation is the ideal approach; however, it is extremely cumbersome and costly. Therefore, we use an LLM-based evaluation to assess the effectiveness of information updates. We utilized average outputs from three closed-source LLMs accessed via their APIs: Gemini 1.5 Flash Pro, GPT-4, and GPT-3.5. \footnote{We avoided open-source models like LLaMA 3.0 due to their weaker performance in semantic similarity tasks, as prior research shows closed-source models excel in capturing subtle semantics.} In our evaluation process, we systematically measured the individual similarity of each aligned row for both {\sc Input-Output} and {\sc Output-Gold} across all three models. After generating prediction scores, we averaged the similarity scores from each model to obtain a consolidated view of overall alignment quality. This ensemble approach proved significantly more effective for similarity matching than relying on a single LLM. By averaging the output of multiple models, we effectively utilized their diverse strengths, enhancing the robustness and accuracy of our semantic evaluations. This also mitigated the limitations associated with any individual model and provided a more consistent semantic matching.

\paragraph{Baselines Methods.} We compare our proposed decomposition approach against multiple baselines, including both \textit{deterministic rule-based} and \textit{generation-based methods}. For the deterministic rule-based comparison, we utilize the rule-based update technique from InfoSync \cite{khincha-etal-2023-infosync}. In the generation-based approach, we take a straightforward approach with direct prompting. We simply describe the task at hand, without breaking it into smaller steps or outlining modeling strategies. The model is then free to tackle the task on its own, using chain-of-thought reasoning to guide its process. Additionally, we adapt the InfoSync technique (Align-Update variants, two and joint prompts), implementing a two-step process involving initial alignment followed by updates with a large language model (LLM). This involves two strategies: one where we provide instructions in a single prompt (joint prompts) to perform both steps simultaneously, and another where we use two sequential prompts—one (two prompts) for alignment and the output of which feeds into a second prompt that handles the final updates. 

\textbf{Ablations.} Additionally, we also compare our approach through various ablation studies, starting with a single prompt containing all instructions, referred to as the Direct Decompose Prompt. We then conduct step-wise ablations of our decomposition method, beginning with just English translation and direct updates, followed by back-translation, which we refer to as Translation (+BackTrans). Next, we incorporate the merge and alignment steps in place of direct updates, referred to as Merge and Alignment. Finally, we transform the output into a knowledge graph before merging and aligning, completing our decomposition methods, which we refer to as Knowledge Graphs. 

\noindent \textbf{Human Baseline.} We also assess our model's performance against a human baseline (from independent research team members), although this evaluation is limited to only 100 randomly selected table updating pairs due to cost and time constraints.

\begin{table*}[!htb]
\centering
\small
\setlength{\tabcolsep}{2pt}  

\begin{tabular}{lc|cc|c|c}
\toprule
                                & Trialign Rows (Tr)          & \multicolumn{2}{c|}{Bialign Rows (Bi)}              & UnAlign Gold (UG) & Input BiAlign (Bi)          \\
\midrule
Methods                         & Updated $\uparrow$  & Added ($\%$) $\uparrow$ &  Added (\#Rows) $\uparrow$ & Missed (G) $\downarrow$ & Delete (I) $\downarrow$\\
\midrule
InfoSync \cite{khincha-etal-2023-infosync}                      & 1.28                     & 12.18                    & 2.99                        & 4.67                     & 0.35                    \\
Direct Prompt                   & 0.63                     & 11.55                    & 3.63                        & 4.40                     & 0.50                    \\
Align-Update (Two Prompts)      & -0.77                    & 12.59                    & 3.98                        & 2.74                     &\bf 0.14                    \\

Align-Update (Joint Prompt)     & 0.51                     & 13.58                    & 3.48                        & 3.24                     & 0.17                    \\
\midrule
\multicolumn{6}{l}{Our Proposed Decomposition Approach} \\
Direct Decompose Prompt                 & 0.90                     & 12.06                    & 2.98                        & 4.65                     & 0.35                    \\

Translation(+BackTrans)               & 0.62                     & 16.88                    & 4.09                        & 3.71                     & 0.38                    \\

+ Merge and Alignment                  & 1.33                     & 17.80                    &\bf 4.99                        & 2.92                     & 0.48                    \\

+ Knowledge Graph                &\bf 1.79                     &\bf 20.58                    & 4.88                        &\bf 2.69                     & 0.45                    \\

\midrule
Human (100 examples)                  & 1.75                     & 21.44                    & 5.6                        & 2.09                     &
\bf 0.12                    \\

\bottomrule
\end{tabular}

\caption{\textbf{Information updation results with average over multiple LLMs.} The performance is reported after using the average of similarity score for multiple LLMs for \textit{information evaluation}, including GPT-3.5, LLaMA 3.0 (70B), and Gemini 1.5 Flash Pro. These results also include ablation studies on various components of our proposed task decomposition, including Translation, KG conversion, and merge-alignment.}
\label{tab:main_all_results}
\vspace{-1.00em}
\end{table*}

\subsection{Results and Analysis}

\textbf{Information Alignment Results.} The results of Alignment are shown in Table \ref{tab:alignment_results}, which show that our multi-voting achieves superior alignment, demonstrating very high precision and recall, with an overall F1 score of 93$\%$. Additionally, multi-voting enhances robustness and reduces variations inherent in each individual model, leading to a more consistent and reliable alignment outcome.

\begin{table}[!htb]
\setlength{\tabcolsep}{4pt}
\centering
\small
\scalebox{0.8}{
\begin{tabular}{p{1.5cm}@{}|lccc}
\toprule
\textbf{Model}                        & \textbf{Type}   & \textbf{Precision} & \textbf{Recall} & \textbf{F1}    \\ \midrule

\multirow{3}{*}{InfoSync} 
                                      & Input\_Gold     & 96.62              & 88.64           & 91.26          \\
                                      & Output\_Gold    & 89.37              & 82.07           & 84.42          \\ 
                                      & Overall Average         & 92.90              & 85.27           & 87.75          \\ \midrule
\multirow{3}{*}{GPT3.5}               
                                      & Input\_Gold     & 96.29              & 93.99           & 94.40          \\
                                      & Output\_Gold    & 89.63              & 85.98           & 86.70          \\ 
                                      & Overall Average        & 92.88              & 89.88           & 90.46          \\\midrule
\multirow{3}{*}{GPT3.5} 
                                      & Input\_Gold     & 98.06              & \bf 94.10           & 95.66          \\
                                      & Output\_Gold    & \bf 94.57              & 87.41           & 89.81          \\ voting(3x) & Overall Average         & 96.27              & 90.67           & 92.66          \\ \midrule
\multirow{3}{*}{Gemini}
                                      & Input\_Gold     & 96.88              & 92.05           & 93.63          \\
                                     & Output\_Gold    & 92.52              & 83.95           & 86.46          \\  voting(3x)  & Overall Average         & 94.65              & 87.90           & 89.96          \\\midrule
\multirow{3}{*}{Multi} 
                                      & Input\_Gold     &\textbf{ 99.15}              & 93.88           & \bf 95.80          \\
                                    & Output\_Gold    & 94.18              & \textbf{88.39}           & \textbf{90.69}          \\   voting(3x) & Overall Average         & \bf 96.60              & \textbf{91.07}           & \textbf{93.18}          \\\bottomrule
\end{tabular}

}
\vspace{-0.5em}
\caption{Evaluation models alignment performance.}
\label{tab:alignment_results}
\vspace{-1.00em}
\end{table}

\noindent \textbf{Information Updation Results.} Our main results, i.e. on updating information, are shown in Table \ref{tab:main_all_results}, demonstrate that our proposed decomposition technique significantly outperforms several baselines, including \citet{khincha-etal-2023-infosync}, particularly in Information Updation (1.79$\%$) and Information Addition (20.58$\%$), highlighting the methods strength in dynamically updating and enriching data.

\begin{table*}[!htb]
\centering
\small
\begin{tabular}{l||c||ccccc}
\toprule
Error Types.	&	In Refer.	&	+Tr. (En)	&	+ KG Cons.	&	+ Merge	&	+ Table Conv.	&	+ Tr. (BT-Orig) \\
\midrule
Missing	&	145	&	145	& \bf	151 (+6)	&\bf	198(+47)	&	\bf 202	(+4)&	202 \\
Outdated (Full)	&	35	&	35	&	35	&\bf	51(+16)	&\bf	59(+8)	&	59 \\
Outdated (Partial)	&	59	&	59	&	59	&\bf	68 (+9)	&\bf	73 (+4)	&	73 \\
Redundant	&	0	&	0	&\bf	66 (+66)	&	66	&	66	&	66 \\ \midrule
Total	&	239	&	239	&\bf	311 (+72)	&\bf\bf	383 (+72)	&\bf	\bf 400 (+17)	&	400 \\
\bottomrule
\end{tabular}

\caption{
\textbf{Error analysis:} Step-wise error analysis of the decomposition pipeline, showing error compounding at each stage: "In Reference" refers to total errors in the input reference tables (\textit{lower bound}), "+Tr. (En)" captures total errors after translation to English, "+KG Cons." indicates total errors after knowledge graph construction, "+Merge" shows total errors after merging, "+Table Conv." tracks total errors after converting graphs to tables, and "+Tr. (BT-Orig)" refers to total errors after back translation. Numbers in parentheses reflect incremental error increases at each stage.}
\vspace{-1.5em}
\label{tab:error-types}
\end{table*}

Our approach excels in correcting erroneous and adding missing information, consistently outperforming other methods. For instance, Align-Update (Two Prompts) shows a negative result (-0.77) in updates, while our approach performs reliably across key metrics. It captures missing data effectively, with the least amount of missed rows (2.69), closest to human performance (2.09 rows). Although there is a slight increase in the deletion rate (0.45) due to more prompts, this is outweighed by the improvements in adding and updating information, demonstrating our method's strength. 

The integration of the \textit{+Knowledge Graph} plays a crucial role in boosting performance, particularly in Information Addition, where we achieve the highest score of 20.58$\%$. The combination of \textit{+merging} and alignment techniques further enhances performance, reflected in the added rows metric (4.99). The model falls slightly behind human performance in deletion rates, but it outperforms in information addition and updating, significantly narrowing gap between human and model.

We observe consistent performance gains across models, with detailed results in Table \ref{tab:main_seperate_results} in Appendix \ref{sec:appendic_individual_results} for GPT 3.5, Gemini 1.5 flash Pro, and LLAMA 3.0 (70B). Our approach consistently outperforms existing methods across architectures. Gemini 1.5 Flash Pro achieves the highest scores, excelling in complex tasks. LLAMA 3.0 performs well but slightly behind Gemini 1.5, whereas GPT-3.5 improves on earlier baselines but lags behind the others. These differences likely reflect variations in architecture and training data, with Gemini’s advanced features providing an edge. Our method’s consistent success across models highlights its broad effectiveness and generalizability.

\textbf{Where do LLM Fail?} We performed a step-wise error analysis of our approach compared to the Gold Tables, classifying the errors into the following categories: \textbf{(a) Missing Information}: A complete row is missing from the table compared to the gold table. \textbf{(b) Outdated Information (Full)}: The entire row contains outdated information when compared to the gold table. \textbf{(c) Outdated Information (Partial)}: Some parts of the row are outdated, while others are up to date. \textbf{(d) Redundant Information}: The model retains redundant data from the input alongside updated information from the reference. This categorization helps identify the specific areas where the model’s performance diverges from the gold standard.

Table \ref{tab:error-types} breaks down the errors introduced at various stages of our decomposition pipeline, categorizing them into four types: Missing Information, Outdated Information (Full), Outdated Information (Partial), and Redundant Information.  Each column tracks errors at different stages: "In Reference" shows baseline errors in the reference tables which are used to update outdated tables. These errors serve as a lower bound because even a perfect method cannot resolve information that is missing from the reference table used for updates. The reference tables initially contain 239 errors, including 145 missing rows, 35 fully outdated rows, and 59 partially outdated rows, with no redundancy.

The subsequent columns detail total errors after each step, i.e. translation to English (+Tr. En), knowledge graph construction (+KG Cons.), merging (+Merge), table conversion, i.e., restructuring (+Table Conv.), and back translation to the original language (+Tr. BT-Orig). The numbers in parentheses show the errors added in that step as compared to the prior step, i.e. new errors introduced in that step. Translation to English adds no new errors. However, errors rise to 400 after the knowledge graph construction and merging stages. Knowledge graph construction introduces 6 missing rows and 66 redundant rows, highlighting issues with removing outdated data. The merging stage adds 47 missing rows, 16 fully outdated rows, and 9 partially outdated rows, indicating challenges with data integration. Improving these stages, particularly knowledge graph construction and merging, could greatly reduce errors and enhance accuracy.

\section{Further Discussion}

\textbf{ Why Zero-Shot over Few-Shot ?} Our proposed method employs hierarchical decomposition to break the problem into multiple, simplified tasks, effectively eliminating the reliance on few-shot learning. Moreover, using a few-shot approach may introduce bias, as the model could perform disproportionately well on categories selected as exemplars. To ensure that information synchronization remains unbiased and generalizable across various entity categories, we opted for a zero-shot setting. Lastly, our preliminary tests revealed that the increased computational and financial costs associated with few-shot learning using GPT APIs resulted in only marginal improvements, making it a less practical choice for our objectives.

\textbf{Why Translation to English?} Translation models for low-resource languages like Afrikaans and Cebuano often struggle with accuracy. Our research shows that direct translations between these languages tend to be unreliable, but translating everything into English boosts performance significantly. To address cultural nuances, we employ a two-way translation strategy. For reverse translations, we include examples from the original English-to-X translations, formatted as X-to-English mappings, as few-shot prompts. This approach helps capture cultural context and ensures more accurate alignment without losing meaning, especially for complex terms and idioms. This trend aligns with the fact that large language models (LLMs) are predominantly trained on English data. Although multilingual models continue to improve, LLMs trained primarily with English datasets still excel at tasks like knowledge graph construction, data merging, and table generation. Translating content into English allows us to leverage these LLMs' capabilities, improving semantic accuracy, consistency, and cross-lingual data processing. Additional discussion on explanation for using Knowledge graphs in the above proposed methods is explained in Appendix section \ref{apdx:sec:importance:KG}, showing an example.

 \section{Related Works}

\paragraph{MultiLingual Information Alignment and Update.}  
Past efforts in multilingual table attribute alignment have employed both supervised and unsupervised techniques. Supervised methods used simple classifiers based on features such as cross-language links and cosine text similarity derived from tables \cite{10.1145/1498759.1498813,zhang2017cross,ta2015model}. On the other hand, unsupervised approaches relied on corpus statistics and template or schema matching for alignment \cite{bouma-etal-2009-cross,nguyen2011multilingual}. Previous research on information updates \cite{iv-etal-2022-fruit,spangher-etal-2022-newsedits,panthaplackel-etal-2022-updated,Zhang:2020:GCS,Zhang:2020:NED} has primarily focused on Wikipedia and news articles, rather than semistructured data like tables. \citet{spangher-etal-2022-newsedits}, specifically, examines the challenge of updating multilingual news articles across different languages. The most closely related work is \cite{khincha-etal-2023-infosync}, which proposed a rule-based approach. However, this method struggles with corner cases and does not leverage current state-of-the-art large language models (LLMs) for multilingual information synchronization. Our work addresses these gaps by introducing an LLM-based prompting approach that is adaptable across different languages, providing a more scalable solution.

\textbf{Temporal Understanding.} Temporal evolving information has been explored through various datasets in the context of question answering. TORQUE \cite{ning-etal-2020-torque} and TIMESENSITIVEQA \cite{chen2021a} focus on time-sensitive questions from Wikipedia, while SYGMA \cite{neelam-etal-2022-sygma}, CRONQUESTIONS \cite{saxena-etal-2021-question}, and TEMPQUESTIONS \cite{10.1145/3184558.3191536} deal with temporal queries in knowledge graphs. SUMIE \cite{hwang2024sumie}, addresses a similar task in a specific domain and shares some similarities with our work. SUMIE deals with textual summarization, analyzing unstructured text. In contrast, our approach focuses on semistructured data, specifically the synchronization of infobox tables. This distinction allows us to tackle a broader range of synchronization tasks that require structured reasoning, beyond just textual content. Our dataset and approach are multilingual, concentrating on the synchronization of tables across different languages—an aspect not covered by SUMIE. We investigate how data can be aligned and updated across multiple languages, whereas SUMIE does not explore multilingual contexts. SUMIE generates data using LLM-based synthetic pipelines, whereas our dataset is directly sourced from real-world Wikipedia data, offering more diversity.

SituatedQA \cite{zhang-choi-2021-situatedqa} and TEMPLAMA \cite{dhingra-etal-2022-time} target open-domain and cloze-style temporal queries. TempTabQA \cite{gupta-etal-2023-temptabqa}, TIQ \cite{10.1145/3589335.3651895}, TRAM \cite{wang2024tram}, and the BIG-bench project \cite{srivastava2023beyond} address temporal reasoning over tables and knowledge bases. More recent work \cite{tan2023benchmarkingimprovingtemporalreasoning, tan2024robusttemporalreasoninglarge} investigates temporal reasoning in large language models (LLMs) using unstructured and synthetic data. 

However, none of this work focuses on editing multilingual tables. Some studies focus on Wikipedia-based document editing \cite{10.1145/1871437.1871698,10.1145/3184558.3191647,10.1145/2396761.2398627}, but not tables. Others apply editing strategies to technical, scientific, legal, and medical tables \cite{wang-etal-2013-transfer,10.1145/3004296}. Expanding our approach to include social, economic, and cultural aspects in table updates would be a valuable direction for future research.

 \section{Conclusion and Future Work}
In this paper, we explored the application of large language models (LLMs) for multilingual information synchronization, focusing on improving the accuracy and coherence of updates to Wikipedia tables in low-resource languages. Our task decomposition strategy significantly outperformed baseline methods, especially in information updating and addition. The Information Updation dataset enabled a more precise evaluation of LLM capabilities. Overall, our findings highlight the potential of LLMs for dynamic data enrichment across diverse architectures, advancing multilingual and low-resource information systems. 

Future research could explore several key directions: (a) extending the dataset to include diverse languages and more complex information structures to test LLM generalizability, (b) integrating LLMs with rule-based methods or knowledge graphs for improved factual accuracy, (c) enhancing the model's performance in deletion tasks without weakening its strength in addition and updating, and (d) investigating efficient prompting strategies and fine-tuning techniques to improve scalability and real-world applicability across different model architectures.
\newpage
\section*{Limitations}
Even though our research demonstrates significant improvements in multilingual information synchronization using large language models (LLMs), several limitations remain. The performance of the models is highly dependent on the quality and diversity of the pre-training data, which may not fully capture the nuances of low-resource languages, leading to inconsistencies across different linguistic contexts, and across different LLMs. Additionally, although our task decomposition strategy improves performance in information updating and addition tasks, it also increases the number of prompts, resulting in a slight rise in deletion errors. This highlights the need for further refinement to balance the model's strengths in information addition and correction with its ability to manage deletions effectively. The use of closed-source models such as GPT-3.5 and Gemini 1.5 Flash Pro also limits transparency and replicability, while open-source models such as LLAMA 3.0 offer more flexibility but may not achieve the same performance levels. Lastly, the computational demands of our approach, though manageable within our project, could pose challenges for broader scalability, particularly in resource-constrained environments. Future research should focus on developing more efficient and scalable solutions to address these limitations and ensure generalizability across diverse languages and domains.

\section*{Ethics Statement}
This research on leveraging large language models (LLMs) for multilingual information synchronization involves several ethical considerations. First, there is a risk of reinforcing biases, particularly in low-resource languages where training data may be limited and skewed, potentially leading to the spread of cultural or factual inaccuracies. Ensuring transparency and incorporating mechanisms for human oversight are essential to prevent misinformation, especially when automating updates for public knowledge sources such as Wikipedia. Additionally, respecting intellectual property and data rights is critical when utilizing publicly available datasets, as unauthorized use could raise ethical and legal concerns. The computational cost of training and deploying LLMs also contributes to environmental impacts, highlighting the importance of developing more energy-efficient models. Although this research demonstrates the potential of LLMs to improve information synchronization, addressing these ethical issues is key to responsible and equitable deployment in real-world applications.

\section*{Acknowledgements}
Research was sponsored by the Army Research Office and was accomplished under Grant Number
W911NF-20-1-0080. The views and conclusions contained in this document are those of the authors and should not be interpreted as representing the official policies, either expressed or implied, of the Army Research Office or the U.S. Government. The U.S. Government is authorized to reproduce and distribute reprints for Government purposes notwithstanding any copyright notation herein. This work was partially funded by ONR Contract N00014-23-1-2364. We extend our gratitude to the annotators who verified our data. We extend our gratitude to the annotators who verified our data extraction and corresponding question answer pairs. We also thank Kashish Ranjan from IIT Guwahati for initial preliminary evaluations of the problem. We extend our sincere appreciation to Jennifer Sheffield from the University of Pennsylvania for her administrative support. Lastly, we extend our appreciation to the reviewing team for their insightful comments.

\bibliography{anthology,custom}
\bibliographystyle{acl_natbib}

\appendix

\section{Appendix}

\subsection{Information Updation Dataset Statistics}
Table \ref{tab:combined_tables} shows the statistics for the proposed information update dataset grouped by different categories and different languages in the table. The dataset is skewed toward high-resource languages because updated tables on entities of interest are seldom available in low-resource languages such as Afrikaans and Cebuano.

\begin{table}[!htb]
\small
    \centering
    \begin{tabular}{lc|lc}
        \toprule
        \textbf{Language} & \textbf{Tables} & \textbf{Category} & \textbf{Tables} \\
        \midrule
        af  & 7    & Album     & 76  \\
        ar  & 120  & Athlete   & 70  \\
        ceb & 4    & City      & 108 \\
        de  & 105  & College   & 112 \\
        en  & 206  & Company   & 148 \\
        es  & 23   & Country   & 122 \\
        fr  & 123  & Musician  & 138 \\
        hi  & 64   & Person    & 108 \\
        ko  & 93   & Stadium   & 66  \\
        nl  & 21   &           &     \\
        ru  & 131  &           &     \\
        sv  & 15   &           &     \\
        tr  & 18   &           &     \\
        zh  & 18   &           &     \\
        \bottomrule
    \end{tabular}
    \caption{\textbf{Dataset Statistics}. Number of pair of (old,new) tables in the dataset grouped the language and different categories. }
    \label{tab:combined_tables}
\end{table}

\subsection{Metrics Definitions for Example in the main paper}
Table \ref{tab:components-associations} shows the tri-align, bi-align and un-aligned set of rows for the example shown in Figure \ref{fig:alignment_diagram} using formal defination from Table \ref{tab:set-theoretic-definitions}. 

\begin{table}[h!]
\small
\centering
\begin{tabular}{p{1.4cm}|p{1.8cm}|p{3.cm}}
\hline
\textbf{Component} & \textbf{Definition} & \textbf{Example Associations} \\ \hline

\textbf{Tri-Align} (\(\texttt{tr}\)) & Alignment across \textbf{Input}, \textbf{Gold}, and \textbf{Output}. & 
\(\texttt{ir}_1 \leftrightarrow \texttt{gr}_1 \leftrightarrow \texttt{or}_1\) \newline
\(\texttt{ir}_2 \leftrightarrow \texttt{gr}_2 \leftrightarrow \texttt{or}_2\) \\ \hline

\textbf{Bi-Align} (\(\texttt{bi}\)) & Alignment between only two components. & 
\textbf{Input $\leftrightarrow$ Gold:} \(\texttt{ir}_3 \leftrightarrow \texttt{gr}_3\) \newline
\textbf{Gold $\leftrightarrow$ Output:} \(\texttt{gr}_4 \leftrightarrow \texttt{or}_3\), \(\texttt{gr}_7 \leftrightarrow \texttt{or}_5\) \\ \hline

\textbf{Un-Aligned} (\(\texttt{un}\)) & Elements not aligned with any component. & 
\textbf{Input:} \(\texttt{ir}_4\) \newline
\textbf{Gold:} \(\texttt{gr}_5, \texttt{gr}_6, \texttt{gr}_8\) \newline
\textbf{Output:} \(\texttt{or}_4, \texttt{or}_6\) \\ \hline

\end{tabular}
 \vspace{-0.5em}
\caption{Summary of Alignment Components and Example Associations for the example in Figure \ref{fig:alignment_diagram}.}
\label{tab:components-associations}
\vspace{-1.0em}
\end{table}

\begin{table}[!ht]
\small
\centering
\begin{tabularx}{\columnwidth}{@{}>{\raggedright\arraybackslash}p{0.15\columnwidth}X@{}}
\toprule
\textbf{Type} & \multicolumn{1}{c}{\textbf{Formal Definition}} \\
\midrule
\multicolumn{2}{c}{\textbf{Trialign}}\\
\bf All & \(\{ (i, o, g) \mid i \in \text{Ti}, o \in \text{To}, g \in \text{Tg}, \text{Aligned}(i,g) \wedge \text{Aligned}(o,g) \}\) \\
\addlinespace
\midrule
\multicolumn{2}{c}{\textbf{Bialign}} \\
\bf Input &
\(\{ i \in \text{Ti} \mid \exists g \in \text{Tg}, \forall o \in \text{To}, \text{Aligned}(i,g) \wedge \neg\text{Aligned}(o,g) \}\) \\
\addlinespace
\bf Output &
\(\{ o \in \text{To} \mid \exists g \in \text{Tg}, \forall i \in \text{Ti}, \text{Aligned}(o,g) \wedge \neg\text{Aligned}(i,g) \}\) \\
\midrule
\addlinespace
\multicolumn{2}{c}{\textbf{UnAlign}} \\
\bf Input &
\(\{ i \in \text{Ti} \mid \forall g \in \text{Tg}, \neg\text{Aligned}(i,g)\}\) \\
\addlinespace
\bf Output &
\(\{ o \in \text{To} \mid \forall g \in \text{Tg}, \neg\text{Aligned}(o,g)\}\) \\
\addlinespace
\bf Gold &
\(\{ g \in \text{Tg} \mid \forall i \in \text{Ti}, \forall o \in \text{To}, \neg\text{Aligned}(i,g) \wedge \neg\text{Aligned}(o,g)\}\) \\
\bottomrule
\end{tabularx}
\vspace{-0.5em}
\caption{Definitions of alignment groups}
\vspace{-0.75em}
\label{tab:set-theoretic-definitions}
\end{table}

Table \ref{tab:notation-metrics-details} shows the metrics defined for each alignment type defined in Section \label{sec:eval_metric} for example presented in Figure \ref{fig:alignment_diagram}. The figure illustrates the calculation of various metrics by presenting an example and evaluating each defined metric step by step.
\begin{table*}[h!]
\small
\centering
\begin{tabular}{p{3cm}|p{5cm}|p{5cm}}
\hline
\textbf{Component} & \textbf{Notation} & \textbf{Information Metrics} \\ \hline

\textbf{Tri-Align} (\(\texttt{tr}\)) & 
\(\text{tr}(o, g)\): Gold elements aligned with Output. \newline
\(\text{tr}(i, g)\): Gold elements aligned with Input. \newline
\(\lvert g \rvert\): Total Gold elements. & 
\textbf{Information Updated:} \newline 
\(\textcolor{blue}{\frac{\text{tr}(o, g) - \text{tr}(i, g)}{\lvert g \rvert} = \frac{2}{8}}\) \newline
Represents the net addition of relevant information to the Output. \\ \hline

\textbf{Bi-Align} (\(\texttt{bi}\)) & 
\(\text{bi}(o, g)\): Gold elements aligned with Output but not Input. \newline
\(\text{bi}(i, g)\): Gold elements aligned with Input but not Output. \newline
\(\lvert g \rvert\): Total Gold elements. & 
\textbf{Noisy Information Added:} \newline
\(\textcolor{green!60!black}{\frac{\text{bi}(o, g)}{\lvert g \rvert} = \frac{2}{8}}\) \newline
Represents misaligned additions from the Output. \newline
\textbf{Noisy Information Deleted:} \newline
\(\textcolor{red}{\frac{\text{bi}(i, g)}{\lvert g \rvert} = \frac{1}{8}}\) \newline
Represents omissions from the Input that were relevant. \\ \hline

\textbf{Un-Aligned} (\(\texttt{un}\)) & 
\(\text{un}(g)\): Unaligned Gold elements. \newline
\(\text{un}(i)\): Unaligned Input elements. \newline
\(\text{un}(o)\): Unaligned Output elements. \newline
\(\lvert g \rvert\): Total Gold elements. \newline
\(\lvert i \rvert\): Total Input elements. \newline
\(\lvert o \rvert\): Total Output elements. & 
\textbf{Missing Information (Gold):} \newline
\(\textcolor{violet}{\frac{\text{un}(g)}{\lvert g \rvert} = \frac{3}{8}}\) \newline
Represents the proportion of relevant Gold elements left unaligned. \newline
\textbf{Noisy Information Added (Input):} \newline
\(\textcolor{brown}{\frac{\text{un}(i)}{\lvert i \rvert} = \frac{1}{4}}\) \newline
Indicates the proportion of irrelevant elements in Input. \newline
\textbf{Noisy Information Added (Output):} \newline
\(\textcolor{teal}{\frac{\text{un}(o)}{\lvert o \rvert} = \frac{2}{6}}\) \newline
Indicates the proportion of irrelevant elements in Output. \\ \hline

\end{tabular}
\caption{Summary of Notation and Metrics for Alignment Components with Information Details}
\label{tab:notation-metrics-details}
\end{table*}

\subsection{Results with Single Model Evaluations} \label{sec:appendic_individual_results}
In the main paper (Table \ref{tab:main_all_results}), we reported results based on voting across multiple model predictions. However, we also evaluated the performance using individual models without voting. These results are presented in Table \ref{tab:main_seperate_results}. It is clearly demonstrated here that our proposed approach outperforms all other approaches discussed consistently across models, further proving its efficacy. From the table we can clearly see that Gemini is the best performing model across most metrics. 

The prompts used to generate outputs are shown in \ref{sec:Prompts}. The evaluation prompt used is shown in \ref{subsec:Evaluation}(it is paired with several examples across multiple languages annotated by us covering a variety of evaluation examples to ensure efficacy).

\subsection{Importance of Using Knowledge Graphs.} \label{apdx:sec:importance:KG}
Our method involves table merging, which often faces challenges due to variations in naming conventions, relationships, and structures across different data sources (in our case table form multilingual pages). To address these issues, the Hierarchical Task Decomposition Prompt utilizes knowledge graphs (KGs), providing a unified structured, hierarchical representation of the data. This structure enables more effective reasoning and merging. For example, let's consider two tables Albert Einstein, current table \ref{tab:Example:KG:1} and outdated table that needs to be updated Table \ref{tab:Example:KG:2}.

\begin{table}[htb]
\small
    \centering
    \begin{tabular}{c|c}
    keys & values \\
    
    \hline
         Name  & Albert Einstein \\
          Birth date & March 14, 1879 \\
           Profession &	Theoretical Physicist  \\
           Country &	Germany, United States  \\
    \end{tabular}
    \caption{\textbf{Current Table} in English.}
    \label{tab:Example:KG:1}
\end{table}

\begin{table}[!htb]
\small
    \centering
    \begin{tabular}{c|c}
    keys & values \\
    \hline
          Name  & Albert Einstein \\
          Birth date & 14 March 1879 \\
           Profession &	 Physicist  \\
           Country &	Germany  \\
    \end{tabular}
    \caption{ \textbf{Outdated Table} which Needs to be updated, translated to English form German.}
    \label{tab:Example:KG:2}
\end{table}

By converting these tables into knowledge graphs, we can align: "Birthdate" and "Date of Birth" as the same entity and "Profession" and "Occupation" as related attributes very easily and accurately. This knowledge graphs representation simplifies merging for the LLM: 
\begin{itemize}
    \item \textbf{Handling Variations in Data Representation}, such as "Birthdate" vs. "Date of Birth" or "Profession" vs. "Occupation". Directly using LLMs on these tables may cause confusion or difficulty in aligning these terms. The LLM might not explicitly recognize that these attributes refer to the same entity. By converting the tables into knowledge graphs, we explicitly capture the relationships between the entities. For example, in a knowledge graph, the entity "Albert Einstein" is connected to "Birthdate", "Profession", and "Country" with clear edges that denote these relationships. Even if different names are used (like "Date of Birth" and "Birthdate"), the LLM can recognize that these refer to the same concept by examining the structure and context in the KG. 
    \item \textbf{Improved Alignment and Merging.} With KGs, the LLM can easily align data across tables based on the semantic relationships represented in the graph. For example: "Birthdate" in Table 1 and "Date of Birth" in Table 2 refer to the same information. "Profession" in Table 1 and "Occupation" in Table 2 are related attributes. Similarly, "Country" and "Nationality" refer to the same concept.
    
    With this graph-based representation, merging the two tables becomes much more straightforward. The graph structure helps resolve ambiguities between different terminologies and aligns the data correctly. The LLM can leverage the hierarchical relationships (e.g., Person → Birthdate → 14 March 1879) to merge the two infoboxes into a unified representation. After converting both tables into knowledge graphs and resolving the semantic mappings, the merged table would look Table \ref{tab:Example:KG:3}.
\begin{table}[!htb]
\small
    \centering
    \begin{tabular}{c|c}
    keys & values \\
    \hline
          Person  & Albert Einstein \\
          Birthdate & 14 March 1879 \\
           Profession/Occupation &	 Theoretical Physicist  \\
           Birth Country &	Germany  \\
           Nationality & Germany, United States \\
    \end{tabular}
    \caption{ \textbf{Merged Table.}}
    \label{tab:Example:KG:3}
\end{table}

    \item \textbf{Improved Reasoning with LLM.} The knowledge graph approach improves performance over directly using LLMs on raw tables for the following reasons: 
    \begin{itemize}
        \item \textit{Hierarchical Reasoning}. The hierarchical nature of KGs enables the LLM to reason more effectively about the relationships between entities and their attributes. This is particularly useful in complex tasks like table merging, where identifying relationships between entities in different tables is crucial.
        \item \textit{Merging Benefit Reasoning.} With KGs, merging data becomes more straightforward because the relationships between entities are explicitly defined. The LLM can merge information by focusing on the nodes and edges that connect related concepts, leading to more accurate integration of data and better reasoning.
Converting tables into knowledge graphs allows the LLM to reason effectively over hierarchical and relational data, handling variations in data representation with greater precision. This approach simplifies tasks like table merging, enabling the LLM to align data, resolve ambiguities, and generate more accurate merged results.
    \end{itemize}
\end{itemize}


\begin{table*}[!htb]
\centering
\small
\scalebox{0.9}{
\setlength{\tabcolsep}{2pt}  

\begin{tabular}{lc|cc|c|c}
\toprule
                                & Trialign Rows (Tr)          & \multicolumn{2}{c|}{Bialign Rows (Bi)} &          UnAlign Gold (UG) & Input BiAlign (Bi)          \\
\midrule
Methods                         & Updated $\uparrow$  & Added ($\%$) $\uparrow$ &  Added (\#Rows) $\uparrow$ & Missed (G) $\downarrow$ & Delete (I) $\downarrow$\\
\midrule
InfoSync \cite{khincha-etal-2023-infosync}                      & 0.94                     & 12.18                    & 2.99                        & 4.67                     & 0.35                    \\ \midrule
\multicolumn{6}{c}{GPT 3.5}\\
Direct Prompt	&	1.34	&	5.44	&	5.14	&	4.03	&	0.43 \\
Align-Update (Two Prompts)	&	-0.37	&	4.45	&	0.88	&	\bf 3.39	&\bf	0.29 \\
Align-Update (Joint Prompt)	&	-0.64	&	1.08	&	0.75	&	5.03	&	0.64 \\
\multicolumn{6}{l}{Our Proposed Decomposition Approach} \\
Direct Decompose Prompt 	&	0.65	&	5.8	&	1.75	&	5.76	&	0.78 \\
Translation(+BackTrans)	&	0.31	&	5.67	&	1.85	&	5.67	&	1.04 \\
+Merge and Alignment	&	0.42	&	8.75	&	2.66	&	4.84	&	1.13 \\
+Knowledge Graph	&	\bf 0.76	&	\bf 12.32	&\bf	3.53	&	3.8	&	1.06 \\
\bottomrule


\multicolumn{6}{c}{Gemini 1.5 Flash Pro}\\
Direct Prompt	&	1.27	&	17.29	&	5.12	&	4.02	&	0.42 \\
Align-Update (Two Prompts)	&	-0.97	&	17.24	&	4.07	&	3.44	&\bf	0.07 \\
Align-Update (Joint Prompt)	&	1.14	&	20.36	&	4.83	&	3.11	&	0.11 \\
\multicolumn{6}{l}{Our Proposed Decomposition Approach} \\
Direct Decompose Prompt 	&	1.04	&	15.67	&	3.63	&	4.03	&	0.12 \\
Translation(+BackTrans)	&	0.59	&	23	&	5.24	&	2.67	&	0.05 \\
+Merge and Alignment	&	1.77	&	22.84	&\bf	6.19	&	\bf 1.91	&	0.16 \\
+Knowledge Graph	&\bf 	2.23	&\bf	25.22	&	5.6	&	2.09	&	0.12 \\
\bottomrule


\multicolumn{6}{c}{LLAMA 3.0 (70B)}\\
Direct Prompt	&	1.25	&	16.27	&	5.01	&	4.14	&	0.42 \\
Align-Update (Two Prompts)	&	-0.97	&	16.09	&	3.99	&	3.53	&\bf	0.06 \\
Align-Update (Joint Prompt)	&	1.04	&	19.29	&	4.75	&	3.22	&	0.12 \\
\multicolumn{6}{l}{Our Proposed Decomposition Approach} \\
Direct Decompose Prompt 	&	1	&	14.72	&	3.55	&	4.15	&	0.13\\
Translation(+BackTrans)	&	0.96	&	21.96	&	5.16	&	2.79	&	0.05\\
+Merge and Alignment	&	1.81	&	21.83	&	6.11	&	2.01	&	0.16\\
+Knowledge Graph	&\bf	2.38	&\bf	24.2	&\bf	5.52	&	\bf 2.18	&	0.17\\
\midrule
Human (100 examples)                  & 1.75                     & 21.44                    & 5.6                        & 2.09                     &
0.12                    \\
\bottomrule
\end{tabular}
}
\caption{ \textbf{Information Updation Results for individual LLMs.} GPT3.5, Gemini 1.5 Flash Pro, LLAMA 3.0 (70B) individual performance.}
\label{tab:main_seperate_results}

\end{table*}
\label{sec:appendix}

\newpage
\section{Prompt Examples}
\label{sec:Prompts}

This section presents example prompts used for hierarchical task decomposition and evaluation in our experiments.

\subsection{Hierarchical Decomposition Prompt}
\subsubsection{Translation(x -> English)}
\begin{tcolorbox}[colframe=blue!50!black, colback=black!10!white, coltitle=black, sharp corners, boxrule=0.8mm, width=\textwidth/2, grow to left by=-1mm, grow to right by=-1mm]
Translate the following 
of Category CATEGORY into English, 
and provide only the translated table as the output.
Ensure that strings with apostrophes are escaped properly using a backslash.
Output table Schema:
[
    ["key","value"],
    ["key","value"]
]

Table:
\end{tcolorbox}
\subsubsection{Table to Knowledge Graph Conversion}
\begin{tcolorbox}[colframe=blue!50!black, colback=black!10!white, coltitle=black, sharp corners, boxrule=0.8mm, width=\textwidth/2, grow to left by=-1mm, grow to right by=-1mm]
Please convert the following table into a knowledge graph and provide the final knowledge graph in a structured json format. 
This table is from the category 
The Output should be in a nested dictionary format.
Ensure you do not miss any information in the original table.

Example Output Knowledge Graph:
\{
  "Person": \{
    "Name": "Karla Camila Cabello Estrabao",
    "Born": "March 3, 1997",
    "Age": "24",
    "Birthplace": "Cojímar, Havana, Cuba"
  \},
  "Occupation": \{
    "Primary": "Singer",
    "Additional": ["Songwriter", "Actress"]
    \}
    ....\}
\end{tcolorbox}

\subsubsection{Knowledge Graph Merge or Alignments}
\begin{tcolorbox}[colframe=blue!50!black, colback=black!10!white, coltitle=black, sharp corners, boxrule=0.8mm, width=\textwidth/2, grow to left by=0mm, grow to right by=0mm]

Given two knowledge graphs containing information about an entity, your task is to merge the graphs while adhering to the following guidelines:

Avoid Duplicate Entries:
Ensure that there are no duplicate nodes and relations in the merged knowledge graph.

Resolve Conflicting Information:
In cases where there is conflicting information for a specific node, use the most updated value to resolve the conflict. If you are still not able to merge the conflict, then prefer the value in Graph B. When you resolve a conflict, only one of the rows should finally be outputted, not both.

Merge Redundant Rows:
Explicitly check for and merge redundant rows holding the same information. Combine them into a single entry, and only one of them should be outputted.

Ultimate Goal:
Create a merged knowledge graph that includes the latest and most accurate information available, without any missing entries. Do not remove any entry during the merging process. Provide only the merged knowledge graph as the output.

Knowledge graphs:
\end{tcolorbox}

\subsubsection{Back-Translation}
\begin{tcolorbox}[colframe=blue!50!black, colback=black!10!white, coltitle=black, sharp corners, boxrule=0.8mm, width=\textwidth/2, grow to left by=-1mm, grow to right by=-1mm]
Convert the knowledge graph into an entity centric table in the format of a list of lists.

Here is an example conversion of knowledge graph A to table A:

Graph A:

Table A:

Now convert Knowledge Graph G to table G following similar keys to table A:

Knowledge Graph G:

Ensure that strings with apostrophes are escaped properly using a backslash.
Output table Schema:
[
    ["key","value"],
    ["key","value"]
    
\end{tcolorbox}

\subsubsection{Translation(English -> x)}
\begin{tcolorbox}[colframe=blue!50!black, colback=black!10!white, coltitle=black, sharp corners, boxrule=0.8mm, width=\textwidth/2, grow to left by=0mm, grow to right by=0mm]
Translate the following English language table of Category 
        Ensure that strings with apostrophes are escaped properly using a backslash.

        Here is an example translation:
        
        Original Table:
        
        Translated Table:

        Now translate the following table:
        
        Output table Schema:
        [
            ["key","value"],
            ["key","value"]
        ]
\end{tcolorbox}

\subsection{Align-Update-Joint}
This prompt is used for Align-Update (Joint Prompts). For Align-Update (Two Prompts), we separate the prompts into two parts: one for alignment and one for the update tasks.
\begin{tcolorbox}[colframe=blue!50!black, colback=black!10!white, coltitle=black, sharp corners, boxrule=0.8mm, width=\textwidth/2, grow to left by=-1mm, grow to right by=-1mm]
Your task is to update Table A(
To help you in the task, you are given a set of alignments. Alignments are a mapping between two tables that match similar information. Use these alignments as a reference to make updates as only aligned rows need to be considered while making updates.
Alignments are in the following format:
[
    ['Table A Aligned Key 1'],['Table B Aligned Key 1'],
    ['Table A Aligned Key 2'],['Table B Aligned Key 2'],
]

Follow these steps:

Identify missing or outdated information in Table A compared to its aligned information in Table B, and update it with the corresponding information from Table B. 
You should add any missing rows present in Table B that are not present in table A. These would be rows of table B that are not present in the set of alignments.
You should also fix any wrong, outdated or missing information present in Table using the set of alignments.
Your solution should ensure that Table A contains complete and accurate information about the entity using data from Table B.
Table A :\\
Table B :\\
Alignments:\\
Provide the updated Table A only in language 
Output table Schema:
[
    ["key","value"],
    ["key","value"]
]
\end{tcolorbox}
\subsection{Direct Decompose}
\begin{tcolorbox}[colframe=blue!50!black, colback=black!10!white, coltitle=black, sharp corners, boxrule=0.8mm, width=\textwidth/2, grow to left by=-1mm, grow to right by=-1mm]
Your task is to update Table A(

Follow these instructions:
1) Translate both tables to English.
2) Create a merged table combining the information from both tables ENSURING that you fix any wrong, outdated or missing information present in both tables.
3) Use the Merged table to update the translated version of Table A.
4) Translate table A back to 

Your solution should ensure that Table A contains complete and accurate information about the entity using data from Table B.
Table A : \\
Table B : \\
Provide the updated Table A ONLY in language 
Ensure that strings with apostrophes are escaped properly using a backslash.
Output table Schema:
[
    ["key","value"],
    ["key","value"]
]
\end{tcolorbox}

\subsection{Just Alignments Prompt}
\begin{tcolorbox}[colframe=blue!50!black, colback=black!10!white, coltitle=black, sharp corners, boxrule=0.8mm, width=\textwidth/2, grow to left by=-1mm, grow to right by=-1mm]
Please provide a list of aligned keys by matching Table G keys with suitable Table A keys, ensuring that they have similar semantic values. 
Allow for multi-alignments where appropriate. If no suitable alignment is found, please skip that key. Do not change the way a key is written and use the exact representation while making alignments.

Tables for Alignment(language):

Table A:

Table G:

Output Schema:
   
    [
        [A-key,G-key],
        [A-key,G-key]....
    ]
\end{tcolorbox}

\clearpage
\subsection{Evaluation Prompt}
\label{subsec:Evaluation}
\begin{tcolorbox}[colframe=blue!50!black, colback=black!10!white, coltitle=black, sharp corners, boxrule=0.8mm, width=\textwidth, grow to left by=0mm, grow to right by=0mm]
Your task involves analyzing two sets of key-value pair tables. Begin by translating the tables to English. Then, extract all pertinent fine-grained details from each table. Then, delve into the semantic content, disregarding minor differences due to formatting, grammar, and language nuances.

Within the tables, information may fall into two categories:

\textbf{`Similar Information'}: Information common to both tables

\begin{itemize}
    \item \textbf{`Consistent Information'}: Both tables contain identical data with possible differences in format.
    \item \textbf{`Contradictory Information'}: Tables present conflicting data with clear difference in meaning.
\end{itemize}

\textit{Note: While analyzing the information, especially for similar information from two tables, solely focus on the semantic content, disregarding any minor differences due to formatting, grammar, and linguistic variations. While comparing numerical information, allow a reasonable error percentage that you consider acceptable before presenting information as inconsistent. Allow the same error margin for other types of information such as coordinates. Be lenient in grouping information as 'Consistent' when slight differences still refer to the overall same data.}

\textbf{`Unique Information'}: Information exclusive to one table.

Your comparison should result in four types of information:

\begin{itemize}
    \item \textbf{Similar and consistent information}: \texttt{similar\_consistent}
    \item \textbf{Similar and contradictory information}: \texttt{similar\_contradictory}
    \item \textbf{Table 1 unique information}: \texttt{table1\_unique}
    \item \textbf{Table 2 unique information}: \texttt{table2\_unique}
\end{itemize}

Here are the test tables provided in language :

\textbf{Table 1:}

\textbf{Table 2:}

\textit{Note: While comparing information, solely focus on the semantic content, disregarding formatting, grammar, and language nuances.}

\end{tcolorbox}
\clearpage









\end{document}